%% file: main.tex
\documentclass[10pt,twocolumn,letterpaper]{article}

\usepackage{wacv}              

\input{preamble}

\definecolor{wacvblue}{rgb}{0.21,0.49,0.74}
\usepackage[pagebackref,breaklinks,colorlinks,allcolors=wacvblue]{hyperref}
\usepackage{comment}
\usepackage{booktabs}
\usepackage{caption}
\usepackage{siunitx}

\def\wacvPaperID{*****} 
\def\confName{WACV}
\def\confYear{2027}

\title{DARA: Degradation-Aware Low-Rank Residual Adaptation with Original-to-Corrupted Distillation for Corruption-Robust Animal Re-Identification}

\author{Cynthia Xie\\
The University of Auckland\\
Auckland, New Zealand\\
{\tt\small zxie211@aucklanduni.ac.nz}
\and
Talia Xu\\
The University of Auckland\\
Auckland, New Zealand\\
{\tt\small talia.xu@auckland.ac.nz}
}

\begin{document}
\maketitle
\input{sec/0_abstract}    
\input{sec/1_intro}
\input{sec/2_related_work}
\input{sec/3_method}
\input{sec/4_experiments}
\FloatBarrier
\input{sec/5_conclusion}
\FloatBarrier
{
    \small
    \bibliographystyle{ieeenat_fullname}
    \bibliography{main}
}
\clearpage
\onecolumn
\appendix
\input{sec/6_appendix}

\end{document}

%% file: preamble.tex
\usepackage{amsfonts}
\usepackage{amsmath}
\usepackage{amssymb}
\usepackage{booktabs}
\usepackage{float}
\usepackage{graphicx}
\usepackage{placeins}
\usepackage{stfloats}



%% file: sec/0_abstract.tex
\vspace{-4mm}
\begin{abstract}
Animal re-identification (Re-ID) relies on fine-grained identity cues that can be disrupted by blur, noise, compression, and other visual degradations. Existing robustness strategies based on degradation-augmented training or pixel-level restoration improve robustness indirectly, but do not explicitly repair shifts in the identity retrieval space. We study corruption-robust animal Re-ID as input-conditioned feature-space repair and introduce DARA, a lightweight retrofit for compact Re-ID models. DARA freezes the fine-tuned backbone and learns routed low-rank residual experts to adapt degraded-input embeddings without corruption-type annotations. To stabilize this adaptive repair, original-to-corrupted distillation uses an original-image teacher to preserve individual embeddings and retrieval relations. Experiments on ATRW, FriesianCattle2017, MPDD, and SeaStarReID2023 show that DARA improves corrupted-query retrieval over standard and augmentation-based fine-tuning, generalizes to unseen corruptions and cross-domain evaluation, and recovers 77.0\% of the corrupted-query mAP gap to full corrupted fine-tuning while adding only 0.49\% parameters and 0.05\% FLOPs.
\end{abstract}

%% file: sec/1_intro.tex
\vspace{-6mm}
\section{Introduction}

Animal re-identification (Re-ID) identifies a specific animal by matching a query image against a gallery of previously observed individuals \cite{ravoor2020survey,cermak2024wildlifedatasets,beyan2026survey}. Because query and gallery images are often captured at different times and under different conditions, an animal must remain recognizable despite changes in appearance. This ability to identify individuals across repeated observations is important for long-term wildlife monitoring and precision livestock management \cite{cermak2024wildlifedatasets,chen2025animalreidmicrocontrollers}. The task is challenging because individuals of the same species may look highly similar, leaving recognition dependent on subtle identity cues that persist across observations. Recent methods become increasingly effective at learning fine-grained representations and now achieve strong performance on standard benchmarks \cite{cermak2024wildlifedatasets,wu2026semanticfeature,beyan2026survey}.

In practical deployments, however, image quality is not always ideal \cite{hendrycks2019benchmarking,polychronou2026degradation}. Sub-optimal imaging conditions can obscure the visual evidence that distinguishes one individual from another. The consequences extend beyond a general loss of image quality: a degraded observation may resemble the wrong individuals more closely and push the correct match lower in the retrieval ranking.
In addition, degradation does not affect every observation in the same way. The same nominal corruption may obscure different identity cues depending on the animal, the image content, and the severity of the degradation. Several forms of degradation may also occur together within a single observation. Therefore, a single fixed correction is unlikely to address this variation effectively.
This input-level variability exposes a limitation of conventional degradation-augmented training. Although such training presents the model with diverse corruptions, it still learns a single set of backbone updates that is applied across all input conditions \cite{polychronou2026degradation}. This full-backbone adaptation is also costly for compact models intended for resource-constrained deployment \cite{chen2025animalreidmicrocontrollers,hu2022lora}.
Static parameter-efficient adapters reduce the number of trainable parameters, but their learned parameters remain input-agnostic after training. They do not explicitly select or combine different corrections according to the degradation characteristics of each image \cite{lou2026parameters,cai2024dynamic}.
Pixel-level restoration offers another possible direction, but producing a visually restored image is not necessarily aligned with recovering the fine-grained identity relationships required for retrieval \cite{yang2023visual,chen2025unirestore,huang2020realworld} . These limitations motivate the following question:
\emph{Can an existing compact Re-ID model be retrofitted with lightweight, corruption-label-free corrections that adapt to each degraded query while preserving the original-image identity geometry learned by the original model?}

We address this question by formulating corruption-robust Re-ID as \emph{input-conditioned repair of corruption-disrupted retrieval geometry}. We introduce DARA, a degradation-aware low-rank residual adaptation framework that keeps the pretrained compact backbone frozen and learns only lightweight residual experts and their router.
The experts provide adaptive \emph{plasticity} by learning complementary low-rank corrections, while the router combines them for each image using degradation-sensitive cues. Neither component relies on predefined corruption categories or corruption-type annotations. Instead, expert specialization emerges from the retrieval objective, allowing the same correction patterns to be reused across degradation conditions.
On the other hand, adaptive repair can distort the identity structure learned from original images. DARA balances this plasticity with \emph{stability} through original-to-corrupted distillation, using a frozen original-image teacher as a structural reference. Individual-level distillation limits embedding drift, while relation-level distillation focuses on the positive and negative pairs most disrupted by corruption. The experts adapt each degraded query, and the teacher anchors the repaired representation to the original-image retrieval geometry.
Together, these components turn DARA into a lightweight robustness retrofit: the backbone remains frozen, and the teacher is removed after training. Across four animal Re-ID datasets, DARA improves corrupted-query retrieval over standard and Aug-Mix fine-tuning while largely preserving performance on unchanged images. Learned routing further improves robustness over single-expert and fixed expert-composition alternatives. It also generalizes to unseen corruptions and domain shift, while adding only 0.49\% parameters and 0.05\% FLOPs.

\begin{itemize}
\item \textbf{Corruption-label-free, input-conditioned feature repair.}
We formulate corruption-robust animal Re-ID as the repair of degradation-induced shifts in identity representations. DARA uses routed low-rank residual experts to adapt the correction to each image while keeping the pretrained backbone frozen, without predefined corruption roles or corruption-type supervision.
\item \textbf{Stability through retrieval-oriented distillation.}
We use the original-image teacher as a structural anchor for adaptive repair. In addition to individual embedding alignment, the proposed relation objective prioritizes positive and negative pairs with the largest teacher--student similarity gaps, directly targeting corruption-induced retrieval errors.
\item \textbf{Robust and parameter-efficient evaluation.}
Across four animal Re-ID datasets, DARA improves corrupted-query retrieval over standard and augmentation-based fine-tuning, generalizes to unseen corruptions and domain shift, and adds only 0.49\% parameters and 0.05\% FLOPs. 
\end{itemize}

%% file: sec/2_related_work.tex
\vspace{-4mm}
\section{Related Work}

\vspace{-1mm}
\subsection{Animal Re-ID}

Animal Re-ID faces stronger fine-grained ambiguity than person Re-ID: individuals of the same species often look highly similar, while the same individual can vary substantially across pose, environment, and capture conditions \cite{ravoor2020survey,hou2025openanimals,beyan2026survey}. This makes animal Re-ID highly dependent on subtle identity cues and motivates animal-specific representation learning \cite{hou2025openanimals,li2025metawild,wu2026semanticfeature}.

Recent work has advanced animal Re-ID through larger datasets, public benchmarks, and stronger representation learning methods \cite{cermak2024wildlifedatasets,adam2025wildlifereid10k,adam2025animalclef,hou2025openanimals,li2025metawild,wu2026semanticfeature}. Other work has addressed specific challenges such as background bias, pose variation, part alignment, and cross-domain generalization \cite{yu2024addressing}. 
In these settings, the identity evidence is usually still visible, even if it is harder to isolate.

Image degradation poses a different challenge: blur, noise, and compression can weaken the identity evidence itself and move a degraded query toward the wrong gallery images in the embedding space. Explicit degradation robustness has only recently been systematically studied in wildlife Re-ID
\cite{polychronou2026degradation}. Our work builds on this direction by studying input-conditioned feature repair, where lightweight corrections are composed separately for each degraded observation.

\vspace{-1mm}
\subsection{Robustness to Image Corruptions}

Image degradation in animal Re-ID is closely related to corruption robustness in broader visual recognition and retrieval. 
Person Re-ID corruption benchmarks show that degrading query or gallery images can substantially reduce matching performance \cite{chen2021benchmarks}. This is especially relevant for animal Re-ID, where retrieval depends on fine-grained identity cues that may be weakened by image corruption.

A common strategy for improving robustness is to train with corrupted samples. In wildlife Re-ID, degradation-based augmented training has been shown to improve retrieval under corrupted inputs \cite{polychronou2026degradation}. Such augmentation strengthens the model through the training distribution rather than changing how each test image is processed. After training, the same learned feature extractor is applied to every query. This differs from input-conditioned feature repair, where the representation can be adjusted according to the degradation affecting a particular observation.

Another related direction is to restore degraded images before recognition. Recognition-driven restoration and unified restoration models increasingly consider downstream task utility rather than perceptual quality alone \cite{yang2023visual,chen2025unirestore}. However, a visually improved image is not necessarily a better Re-ID input: low-level enhancement may change or smooth out subtle identity cues, so improving pixel quality does not necessarily preserve the relationships needed to match the same individual across query and gallery images \cite{huang2020realworld}. DARA instead targets feature-space robustness directly, adapting retrieval representations without requiring an image-restoration stage.

\subsection{Parameter-Efficient Adaptation}

Parameter-efficient adaptation offers a lightweight way to improve robustness while keeping most of the Re-ID backbone unchanged. Adapters and LoRA reduce trainable parameters by adding small modules or low-rank updates \cite{houlsby2019parameter,hu2022lora,hayou2024loraplus,xin2024parameter}. However, most parameter-efficient modules are fixed after training and apply the same learned update to every input. This is limiting when test-time changes are input-dependent, as in image degradation, where different observations may require different feature corrections. This motivates adaptive parameter-efficient modules whose updates can vary with the input \cite{lou2026parameters,cai2024dynamic}.

Adaptive parameter-efficient modules have also been explored in Re-ID and broader vision tasks. In Re-ID, AdalReID \cite{qian2024adalreid} and SAGE-reID \cite{nehdi2026lowrank} organize adapters around dataset-, domain-, or source-specific knowledge for person Re-ID adaptation. Beyond Re-ID, RobuMTL \cite{shaffee2026robumtl} routes LoRA experts using predefined perturbation types and perturbation-label supervision, while Parameters-as-Experts \cite{lou2026parameters} studies low-rank expert composition for general vision adaptation. These methods show the value of adaptive PEFT, but they do not target input-specific repair of corruption-induced shifts in a query--gallery identity retrieval space.

DARA instead uses parameter-efficient routing to repair corrupted Re-ID representations. Its residual experts learn correction directions for degraded animal images, and the router combines them separately for each input without corruption-type supervision. Thus, the adapters are not only a lightweight substitute for full fine-tuning; they provide input-conditioned corrections for corrupted identity representations.

\vspace{-1mm}
\subsection{Original-to-Corrupted Distillation}

Knowledge distillation transfers information from a teacher to a student model \cite{hinton2015distilling} and has been used in Re-ID for scalable person Re-ID, multiple-view Re-ID, and occluded person Re-ID \cite{wu2019distilled,porrello2020robust,zheng2021pose}. For corruption-robust Re-ID, an original-image teacher is useful because it provides the embedding structure before degradation is applied. This matters because corruptions can change nearest-neighbor rankings even when identity labels remain unchanged. Relational distillation also shows that sample relationships can provide supervision beyond individual predictions \cite{park2019relational}.

Recent corruption-robust Re-ID methods use teacher-based alignment or structural consistency to reduce degradation-induced feature shifts. Zhang et al. \cite{zhang2025corruptioninvariant} align corrupted person features with teacher representations from original inputs, while Zhao et al. \cite{zhao2026mixeddegradation} reduce original-degraded distribution discrepancies and preserve structural relations under mixed degradations. These methods show the value of aligning degraded inputs with an original-image reference, but their supervision is not centered on the retrieval pairs most affected by corruption. In particular, they do not explicitly prioritize the positive and negative pairs whose similarities change the most under degradation.

DARA uses original-to-corrupted distillation as a retrieval-oriented constraint. It limits the shift of each corrupted embedding and also supervises pairwise relations. The relation-level loss mines positive and negative pairs with the largest teacher--student similarity gaps, so supervision is concentrated where corruption most changes the retrieval structure. This complements the routed residual experts: the experts adapt each degraded representation, while distillation keeps the repaired embeddings aligned with the original-image identity space.

%% file: sec/3_method.tex
\begin{figure*}[!t]
  \centering
  \includegraphics[width=0.96\textwidth]{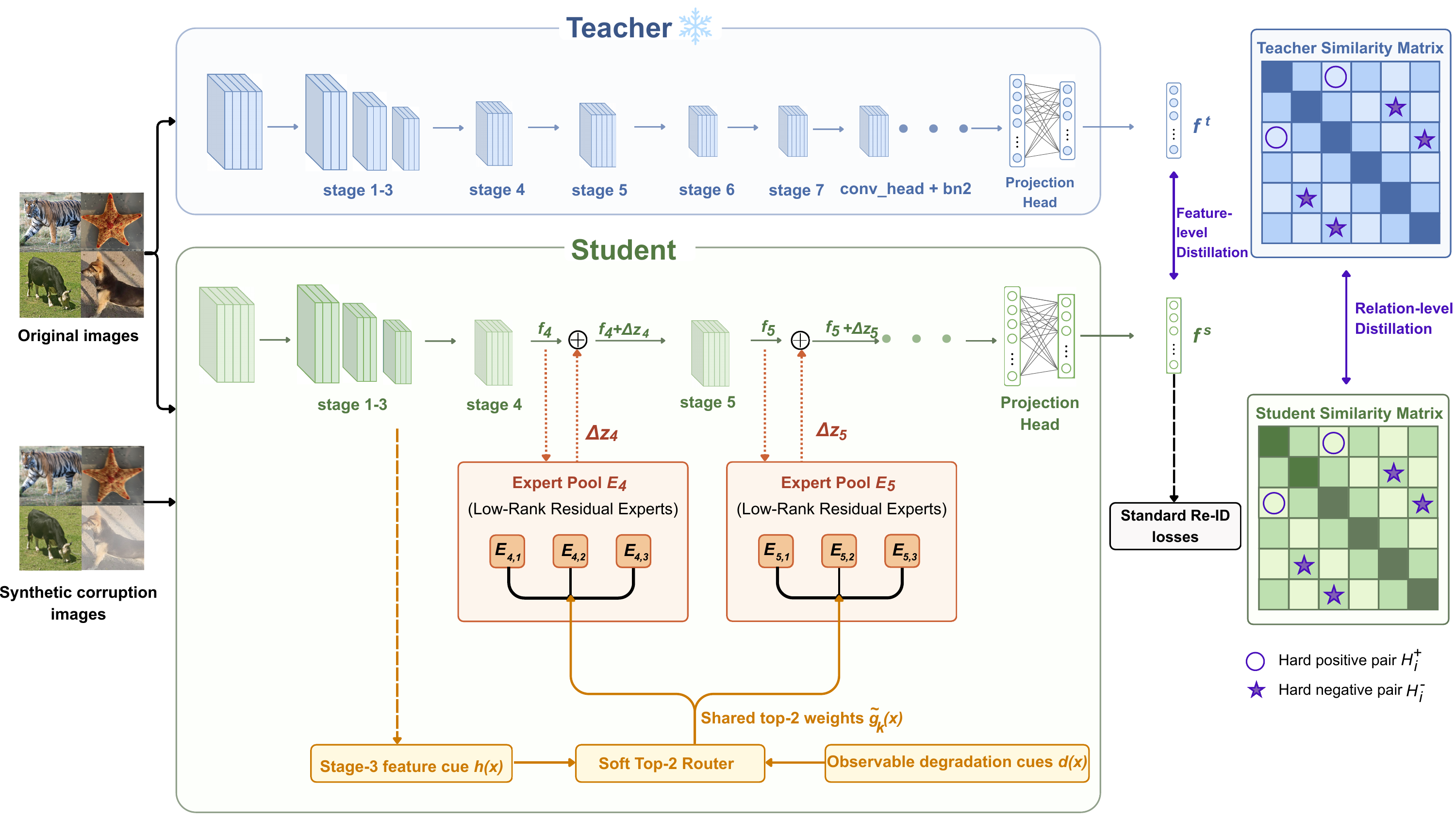}
  \caption{
  Overall framework of DARA for corruption-robust animal Re-ID.
  During training, each original image is paired with a corrupted view. A frozen
  teacher provides original-image references, while the student uses
  degradation-aware low-rank residual experts on a staged compact Re-ID backbone
  to adapt its embeddings. Original-to-corrupted distillation preserves both individual embeddings and retrieval relations, 
  together with standard Re-ID supervision on the student. At inference time, only the adapted student is used.}
  
  \label{fig:framework}
\end{figure*}

\vspace{-2mm}
\section{Method}

DARA starts from a Re-ID model trained on original images. A frozen copy serves as the teacher, while the student is initialized from the same checkpoint and augmented with residual experts and a router. As shown in Fig.~\ref{fig:framework}, each training image is paired with a corrupted view: the teacher embeds the original image, and the student embeds the corrupted view for distillation. The student is also trained with standard Re-ID losses on both original and corrupted views. Routed low-rank residual experts are inserted at stages 4 and 5 (\autoref{sec:dara_experts}), and original-to-corrupted distillation preserves both individual embeddings and pairwise relations (\autoref{sec:dara_distill}). \autoref{sec:dara_opt} gives the full objective and inference procedure, where the teacher is removed and only the adapted student is used for retrieval.

We consider the corrupted-query retrieval setting, where degradations are applied to query images while gallery images remain unchanged. This reflects a common Re-ID scenario: a newly captured field observation may be blurred, noisy, or compressed, but it must still retrieve unchanged reference images of the same individual from an existing gallery. DARA therefore treats robustness as label-free feature-space repair: the goal is to recover identity-discriminative embeddings from degraded inputs while preserving compatibility with the original-image retrieval space of the frozen backbone.

\vspace{-1mm}
\subsection{Degradation-Aware Low-Rank Experts}
\label{sec:dara_experts}

This subsection describes the student-side adaptation modules in Fig.~\ref{fig:framework}: the stage-4 and stage-5 residual expert banks and the router that composes them for each input. Each expert predicts a compact channel-wise correction, while the router selects and weights experts using observable degradation cues and intermediate feature context.

\vspace{-2mm}
\paragraph{Feature-space residual adaptation.}
At each adapted feature stage, the student applies a low-rank residual correction to the intermediate feature map. Given the feature map
$F_l \in \mathbb{R}^{B \times C_l \times H_l \times W_l}$ at stage $l$, we
first obtain a channel descriptor by global average pooling,
$z_l=\operatorname{GAP}(F_l) \in \mathbb{R}^{B \times C_l}$. A low-rank
residual expert then predicts a channel-wise correction:
\[
\Delta z_l = z_l A_l^\top B_l^\top,
\]
where $A_l \in \mathbb{R}^{r_l \times C_l}$ and
$B_l \in \mathbb{R}^{C_l \times r_l}$ are the down-projection and up-projection
matrices, with $r_l \ll C_l$. This factorization replaces a full
$C_l \times C_l$ mapping with a compact low-rank transformation, thereby
limiting the number of additional trainable parameters.

For feature-map stages, the predicted residual is broadcast to the spatial
dimensions and added to the original feature map:
\[
\tilde{F}_l=F_l+\operatorname{Broadcast}(\Delta z_l).
\]
The corrected feature $\tilde{F}_l$ replaces $F_l$ as the input to the next
backbone stage.

\vspace{-1mm}
\paragraph{Degradation-aware expert composition.}
To adapt the residual correction to each degraded input, we use a router to compose a small bank of low-rank experts. Each adapted stage contains an expert pool of size $M=3$. We denote the expert pool at stage $l$ as $\mathcal{E}_l=\{E_{l,1},\ldots,E_{l,M}\}$. The router predicts one shared expert-weight vector per input and reuses it across the adapted stages, while the expert parameters remain stage-specific. The router and experts are trained jointly with the Re-ID and distillation objectives, without corruption-type labels or predefined expert categories.

The router produces a score $g_k(x)$ for each expert index
$k \in \{1,\ldots,M\}$. We first select the two experts with the largest
router scores:
\[
\mathcal{T}(x)
=
\operatorname{Top2}_{k \in \{1,\ldots,M\}}(g_k(x)).
\]
The selected scores are then normalized with a softmax over the selected
experts:
\[
\tilde{g}_{k}(x)
=
\frac{\exp(g_k(x))}
{\sum_{j \in \mathcal{T}(x)} \exp(g_j(x))},
\quad k \in \mathcal{T}(x).
\]
The final residual update is computed as
\[
\Delta z_l
=
\sum_{k \in \mathcal{T}(x)}
\tilde{g}_{k}(x)\Delta z_l^{k},
\]
where $\Delta z_l^{k}$ is the residual predicted by expert $k$.

The router input consists of two cues. The first is an observable degradation cue $d(x)$: a z-score normalized 9-dimensional statistics vector summarizing sharpness, intensity and exposure, frequency content, and compression artifacts. Specifically, it contains Laplacian energy, saturated-pixel ratio, dark-pixel ratio, grayscale mean, RMS contrast, high-frequency FFT ratio, dark-channel response, mean saturation, and block-jump score. The second is a learned 16-dimensional stage-3 feature cue $h(x)$ projected from the stage-3 feature descriptor. The concatenated 25-dimensional cue is passed to a two-layer router MLP with hidden dimension 32 and output dimension $M=3$.

\subsection{Original-to-Corrupted Distillation}
\label{sec:dara_distill}

This subsection describes the teacher-student supervision on the right side of Fig.~\ref{fig:framework}: individual embedding alignment and relation-level alignment between teacher and student similarity structures. Identity labels remain unchanged under corruption, but the relative positions of samples in the embedding space can shift, changing nearest-neighbor rankings. Original-to-corrupted distillation anchors the adapted student to the original-image teacher.

For each original image $x_i$ and corrupted view $\hat{x}_i$, the teacher extracts a reference embedding $f_i^t=T(x_i)$, while the student extracts a corrupted-input embedding $f_i^s=S(\hat{x}_i)$. Both $f_i^t$ and $f_i^s$ denote the 512-D Re-ID embeddings after the projection BN neck, which are also used to compute query-gallery similarities during evaluation.

\vspace{-1mm}
\paragraph{Individual-level distillation.}
The individual loss aligns each corrupted-input student embedding with the corresponding original-image teacher embedding:
\[
\mathcal{L}_{\mathrm{ind}}
= \frac{1}{B}\sum_{i=1}^{B}
\left(1-
\bar{f}_i^{s\top}\bar{f}_i^t
\right).
\]
Here, $\bar{f}$ denotes the $\ell_2$-normalized embedding.
Unlike standard logit distillation, this loss acts directly on the final Re-ID
embedding used for nearest-neighbor retrieval. The frozen teacher provides an
original-image reference for each corrupted image, while the student learns
degradation-adaptive residual corrections.

\vspace{-1mm}
\paragraph{Relation-level distillation.}
Individual alignment corrects each sample independently, but Re-ID depends on the relative structure among samples. A corrupted embedding may remain close to its original counterpart while still altering the ranking order among visually similar identities. To preserve retrieval geometry, we further distill hard pairwise relations. Let
\[
s_{ij}^t=\bar{f}_i^{t\top}\bar{f}_j^t,\qquad
s_{ij}^s=\bar{f}_i^{s\top}\bar{f}_j^s
\]
denote the teacher and student cosine similarities for a pair of samples. For
an anchor $i$, we define the teacher-student relation gap as
\[
d_{ij}=|s_{ij}^t-s_{ij}^s|.
\]
Instead of matching all pairwise similarities in the mini-batch, we mine the
relations whose teacher structure from original images is most poorly reproduced by the corrupted-input student. Specifically, from the positive set $\mathcal{P}_i=\{j:y_j=y_i,j\neq i\}$ and negative set $\mathcal{N}_i=\{j:y_j\neq y_i\}$, we set $K=8$ and select up to $K$ hard positive pairs and up to $K$ hard negative pairs with the largest $d_{ij}$. Mini-batches use identity-balanced sampling with batch size 64 and four images per identity. When fewer than $K$ valid pairs exist, all are used.
The hard sets are defined as
\[
\mathcal{H}_i^+
=\{(i,j)\mid j\in\operatorname{TopK}_{j'\in\mathcal{P}_i}^{K}(d_{ij'})\},
\]
\[
\mathcal{H}_i^-
=\{(i,j)\mid j\in\operatorname{TopK}_{j'\in\mathcal{N}_i}^{K}(d_{ij'})\}.
\]
Here, $\operatorname{TopK}$ returns the indices associated with the largest
relation gaps.
Anchors without valid positive or negative pairs in the mini-batch are skipped.
The relation-level distillation loss is then
\[
\mathcal{L}_{\mathrm{rel}}
=\frac{1}{|\mathcal{H}|}
\sum_{(i,j)\in\mathcal{H}}
\left(s_{ij}^s-s_{ij}^t\right)^2,
\quad
\mathcal{H}=\bigcup_i(\mathcal{H}_i^+\cup\mathcal{H}_i^-).
\]
By focusing on positive and negative pairs with the largest teacher-student
relation gaps, this loss preserves the most corruption-sensitive retrieval
relations while avoiding easy pairs that are already well aligned.

\subsection{Optimization and Inference}
\label{sec:dara_opt}

The student is optimized with the standard Re-ID objective and the two distillation losses:

\[
\mathcal{L}
=\mathcal{L}_{\mathrm{reid}}
+\lambda_{\mathrm{ind}}\mathcal{L}_{\mathrm{ind}}
+\lambda_{\mathrm{rel}}\mathcal{L}_{\mathrm{rel}},
\]
where $\mathcal{L}_{\mathrm{reid}}$ denotes the identity classification and
triplet losses.The Re-ID loss is applied to both the original and corrupted views processed by the student, while distillation uses the teacher embedding from the original image as the original-image reference. The coefficients $\lambda_{\mathrm{ind}}$ and
$\lambda_{\mathrm{rel}}$ balance the individual-level and relation-level
distillation terms. During training, the teacher and the
student backbone weights are frozen. The trainable parameters are the low-rank
residual experts and the router.

At inference time, the adapted student uses the same soft top-2 router to
extract query and gallery embeddings from image statistics and feature cues,
without requiring corruption labels. Although gallery images remain unchanged in the corrupted-query protocol, they are still encoded by the adapted student. This ensures that query and gallery embeddings are produced by the same inference model after the teacher is removed.

%% file: sec/4_experiments.tex
\section{Experiments}

\subsection{Experimental Setting}

\subsubsection{Datasets}

We evaluate DARA on four animal re-identification datasets: ATRW~\cite{li2019atrw}, FriesianCattle2017~\cite{andrew2017visual}, MPDD~\cite{he2023animal}, and SeaStarReID2023~\cite{wahltinez2024opensource}. These datasets cover livestock and wildlife Re-ID across different species, dataset scales, and identity cues. For each dataset, we train on the training split and evaluate using the standard query-gallery retrieval protocol. Dataset statistics are provided in the supplementary material.

\subsubsection{Degradation Evaluation Protocol}

We use corrupted-query (CQ) retrieval as the main robustness setting. Synthetic corruptions are applied only to query images, while gallery images remain unchanged. This setting reflects practical animal Re-ID deployments, where a degraded field image is used as the query and matched against original reference images in the gallery. Following prior corruption-robust Re-ID evaluation \cite{chen2021benchmarks}, we introduce corrupted queries using eight ImageNet-C operators~\cite{hendrycks2019benchmarking}: Gaussian noise, motion blur, de-focus blur, fog, brightness, contrast, JPEG compression, and pixelate. For each query image, we randomly sample one corruption type and one severity level from ${1,2,3,4,5}$. We repeat this process for 10 trials and report the average result. We also report unchanged-image retrieval to measure standard Re-ID preservation. Corrupted-gallery and fully corrupted query-gallery evaluations are provided in the supplementary material.

\vspace{-1mm}
\subsubsection{Evaluation Metrics}

We report mAP, Rank-1, Rank-5, and mINP. CQ mAP and CQ mINP are the primary robustness metrics, while standard mAP measures unchanged-image retrieval preservation. mINP is included because it emphasizes the hardest correct match and is therefore useful for evaluating whether difficult identity relationships are preserved under degradation.

\vspace{-1mm}
\subsubsection{Implementation Details}

We use MobileNetV2~\cite{sandler2018mobilenetv2} as the backbone for all methods. Std-ZS denotes zero-shot evaluation with the ImageNet-pretrained backbone. Std-FT denotes full fine-tuning on original training images without synthetic corruptions. AugMix-FT applies AugMix during full-parameter fine-tuning, while Corrupt-FT performs full-parameter fine-tuning using synthetically corrupted training images. Corrupt-FT therefore serves as a full-backbone corrupted-training baseline.

DARA starts from the Std-FT checkpoint. A frozen copy of this checkpoint is used as the teacher, and the student is initialized from the same checkpoint. During adaptation, the backbone is frozen and only the residual experts and router are optimized. Unless otherwise specified, DARA uses two adapted stages, soft top-2 routing, three residual experts, 50 training epochs, batch size 64, and input size $256 \times 128$. At inference time, the teacher branch and distillation losses are removed, and only the adapted student is deployed.

\subsection{Experimental Results}

We first compare DARA with standard and corruption-aware training baselines, then analyze the contributions of residual repair, distillation, and routing. We also test whether the learned repair generalizes to unseen corruptions and cross-domain data, and measure its deployment overhead.

\subsubsection{Main Results}

Table~\ref{tab:main_results} compares MobileNetV2-based methods under standard and corrupted-query settings. Standard fine-tuning preserves unchanged-image retrieval, but its CQ mAP drops to 30.17 on average, showing that original-image training alone does not provide reliable retrieval under degradation. AugMix-FT improves average CQ mAP to 41.67, while DARA further increases it to 55.70 and improves CQ mAP on all four datasets. At the same time, DARA maintains standard mAP close to the original fine-tuned model, indicating that the learned repair improves robustness without sacrificing unchanged-image retrieval.

Compared with Corrupt-FT, which updates the full backbone using corrupted training images, DARA operates under a stricter retrofit setting: the backbone is frozen and only the residual experts and router are learned. Even with this constraint, DARA recovers approximately 77\% of the CQ mAP gap between Std-FT and Corrupt-FT. This shows that most of the robustness benefit of full corrupted fine-tuning can be recovered through lightweight feature-space repair, while preserving the original compact backbone.

\begin{table*}[t]
  \centering
  \footnotesize
  \caption{Main results under standard and corrupted-query settings.}
  \label{tab:main_results}
  \begin{tabular}{llccccccc}
    \toprule
    Dataset & Method
    & \multicolumn{3}{c}{Standard}
    & \multicolumn{4}{c}{Corrupted Query} \\
    \cmidrule(lr){3-5} \cmidrule(lr){6-9}
    & & mAP & R1 & R5 & mAP & R1 & R5 & mINP \\
    \midrule
    ATRW & Std-ZS & 48.10 & 92.00 & 96.90 & 21.99 & 37.95 & 57.22 & 6.93 \\
    ATRW & Std-FT & 52.00 & 92.20 & 98.10 & 17.37 & 24.17 & 39.91 & 8.30 \\
    ATRW & AugMix-FT & 55.90 & 96.50 & 99.50 & 29.51 & 46.16 & 62.83 & 11.36 \\
    ATRW & Corrupt-FT & 52.40 & 94.60 & 99.10 & 48.82 & 90.09 & 96.96 & 16.65 \\
    ATRW & DARA & 54.20 & 94.30 & 98.80 & 43.28 & 75.33 & 88.96 & 15.91 \\
    \midrule
    FriesianCattle2017 & Std-ZS & 77.40 & 94.10 & 98.80 & 50.10 & 55.76 & 78.12 & 31.70 \\
    FriesianCattle2017 & Std-FT & 95.10 & 98.80 & 100.00 & 55.91 & 53.88 & 72.00 & 45.14 \\
    FriesianCattle2017 & AugMix-FT & 97.10 & 100.00 & 100.00 & 65.18 & 64.12 & 77.76 & 56.03 \\
    FriesianCattle2017 & Corrupt-FT & 94.90 & 97.60 & 100.00 & 90.57 & 96.47 & 99.88 & 77.99 \\
    FriesianCattle2017 & DARA & 95.90 & 100.00 & 100.00 & 85.07 & 90.12 & 97.65 & 72.17 \\
    \midrule
    MPDD & Std-ZS & 69.30 & 85.60 & 91.30 & 32.23 & 37.50 & 54.04 & 20.04 \\
    MPDD & Std-FT & 72.90 & 84.60 & 94.20 & 25.50 & 29.81 & 44.04 & 14.72 \\
    MPDD & AugMix-FT & 79.10 & 91.30 & 97.10 & 41.64 & 49.33 & 64.71 & 28.39 \\
    MPDD & Corrupt-FT & 74.70 & 88.50 & 98.10 & 68.25 & 82.88 & 94.81 & 47.23 \\
    MPDD & DARA & 73.20 & 85.60 & 93.30 & 56.92 & 68.85 & 82.79 & 36.88 \\
    \midrule
    SeaStarReID2023 & Std-ZS & 24.40 & 43.90 & 78.30 & 15.53 & 16.56 & 41.32 & 9.41 \\
    SeaStarReID2023 & Std-FT & 55.20 & 69.30 & 93.10 & 21.88 & 19.42 & 41.80 & 15.27 \\
    SeaStarReID2023 & AugMix-FT & 67.10 & 81.50 & 95.20 & 30.34 & 29.15 & 50.05 & 20.52 \\
    SeaStarReID2023 & Corrupt-FT & 54.10 & 67.20 & 89.90 & 45.76 & 55.08 & 83.49 & 29.82 \\
    SeaStarReID2023 & DARA & 53.50 & 74.60 & 91.00 & 37.53 & 41.90 & 73.39 & 24.93 \\
    \bottomrule
  \end{tabular}
\end{table*}

\begin{table}[t]
\centering
\footnotesize
\caption{Average robustness--retrofit trade-off across four datasets.}
\label{tab:avg_tradeoff}
\setlength{\tabcolsep}{3pt}
\begin{tabular}{lcccc}
\toprule
Method & Adaptation Scope & Extra Params & CQ mAP & Gap Rec. \\
\midrule
Std-FT & Full backbone &  0.00\% & 30.17 & 0.0\% \\
AugMix-FT & Full backbone &  0.00\% & 41.67 & 34.7\% \\
Corrupt-FT & Full backbone &  0.00\% & 63.35 & 100.0\% \\
DARA & Experts + router & +0.49\% & 55.70 & 77.0\% \\
\bottomrule
\end{tabular}
\end{table}

\begin{table}[t]
\centering
\scriptsize
\caption{Progressive component ablation.}
\label{tab:progressive_ablation}
\setlength{\tabcolsep}{1.0pt}
\resizebox{\linewidth}{!}{%
  \begin{tabular}{llccccccc}
    \toprule
    Dataset & Variant
    & \multicolumn{3}{c}{Standard}
    & \multicolumn{4}{c}{Corrupted Query} \\
    \cmidrule(lr){3-5} \cmidrule(lr){6-9}
    & & mAP & R1 & R5 & mAP & R1 & R5 & mINP \\
    \midrule
    ATRW & Std-FT & 52.00 & 92.20 & 98.10 & 17.37 & 24.17 & 39.91 & 8.30 \\
    ATRW & 1-layer & 53.40 & 94.10 & 98.60 & 41.48 & 73.61 & 87.67 & 14.01 \\
    ATRW & 2-layer & 54.20 & 95.00 & 98.10 & 42.77 & 75.61 & 89.95 & 14.64 \\
    ATRW & 2+O2C & 54.20 & 94.30 & 98.80 & 43.28 & 75.33 & 88.96 & 15.91 \\
    \midrule
    Friesian & Std-FT & 95.10 & 98.80 & 100.00 & 55.91 & 53.88 & 72.00 & 45.14 \\
    Friesian & 1-layer & 91.90 & 100.00 & 100.00 & 74.30 & 81.53 & 90.94 & 60.73 \\
    Friesian & 2-layer & 92.80 & 100.00 & 100.00 & 79.14 & 85.18 & 95.41 & 64.13 \\
    Friesian & 2+O2C & 95.90 & 100.00 & 100.00 & 85.07 & 90.12 & 97.65 & 72.17 \\
    \midrule
    MPDD & Std-FT & 72.90 & 84.60 & 94.20 & 25.50 & 29.81 & 44.04 & 14.72 \\
    MPDD & 1-layer & 69.40 & 84.60 & 94.20 & 47.00 & 60.58 & 76.92 & 27.24 \\
    MPDD & 2-layer & 71.10 & 84.60 & 93.30 & 48.43 & 60.29 & 77.60 & 29.14 \\
    MPDD & 2+O2C & 73.20 & 85.60 & 93.30 & 56.92 & 68.85 & 82.79 & 36.88 \\
    \midrule
    SeaStar & Std-FT & 55.20 & 69.30 & 93.10 & 21.88 & 19.42 & 41.80 & 15.27 \\
    SeaStar & 1-layer & 51.00 & 67.20 & 93.10 & 29.22 & 31.69 & 65.87 & 16.58 \\
    SeaStar & 2-layer & 52.00 & 67.70 & 87.80 & 30.90 & 32.17 & 64.60 & 18.95 \\
    SeaStar & 2+O2C & 53.50 & 74.60 & 91.00 & 37.53 & 41.90 & 73.39 & 24.93 \\
    \bottomrule
  \end{tabular}
}
\end{table}

\begin{table}[t]
\centering
\scriptsize
\caption{Expert composition and learned routing ablation.}
\label{tab:router_ablation}
\setlength{\tabcolsep}{1.0pt}
\resizebox{\linewidth}{!}{%
  \begin{tabular}{llccccccc}
    \toprule
    Dataset & Variant
    & \multicolumn{3}{c}{Standard}
    & \multicolumn{4}{c}{Corrupted Query} \\
    \cmidrule(lr){3-5} \cmidrule(lr){6-9}
    & & mAP & R1 & R5 & mAP & R1 & R5 & mINP \\
    \midrule
    ATRW & Single & 53.90 & 93.90 & 98.10 & 41.22 & 74.03 & 88.54 & 14.34 \\
    ATRW & Fixed-2 & 53.80 & 93.20 & 97.40 & 42.76 & 74.74 & 88.61 & 15.46 \\
    ATRW & Fixed-3 & 53.70 & 92.70 & 98.10 & 42.54 & 74.83 & 89.08 & 15.30 \\
    ATRW & Top-2 (3) & 54.20 & 94.30 & 98.80 & 43.28 & 75.33 & 88.96 & 15.91 \\
    ATRW & Top-2 (4) & 54.00 & 95.00 & 98.80 & 43.35 & 76.42 & 89.91 & 15.80 \\
    \midrule
    Friesian & Single & 95.10 & 100.00 & 100.00 & 81.26 & 85.53 & 96.00 & 68.65 \\
    Friesian & Fixed-2 & 95.50 & 100.00 & 100.00 & 82.11 & 87.29 & 95.76 & 70.45 \\
    Friesian & Fixed-3 & 95.40 & 100.00 & 100.00 & 82.03 & 86.82 & 95.65 & 70.29 \\
    Friesian & Top-2 (3) & 95.90 & 100.00 & 100.00 & 85.07 & 90.12 & 97.65 & 72.17 \\
    Friesian & Top-2 (4) & 95.80 & 98.80 & 100.00 & 84.56 & 89.29 & 96.94 & 72.71 \\
    \midrule
    MPDD & Single & 73.20 & 85.60 & 92.30 & 50.00 & 59.58 & 78.33 & 31.44 \\
    MPDD & Fixed-2 & 74.00 & 84.50 & 91.30 & 51.75 & 63.46 & 80.67 & 32.75 \\
    MPDD & Fixed-3 & 73.70 & 85.50 & 92.30 & 51.54 & 62.40 & 80.48 & 32.69 \\
    MPDD & Top-2 (3) & 73.20 & 85.60 & 93.30 & 56.92 & 68.85 & 82.79 & 36.88 \\
    MPDD & Top-2 (4) & 73.40 & 83.70 & 93.30 & 54.74 & 65.96 & 80.87 & 34.51 \\
    \midrule
    SeaStar & Single & 53.90 & 73.52 & 90.82 & 34.76 & 37.52 & 68.05 & 23.47 \\
    SeaStar & Fixed-2 & 53.50 & 74.17 & 90.50 & 35.83 & 38.20 & 70.48 & 24.30 \\
    SeaStar & Fixed-3 & 53.90 & 74.22 & 91.92 & 36.51 & 39.58 & 71.43 & 24.74 \\
    SeaStar & Top-2 (3) & 53.50 & 74.60 & 91.00 & 37.53 & 41.90 & 73.39 & 24.93 \\
    SeaStar & Top-2 (4) & 53.90 & 74.10 & 91.10 & 36.04 & 38.52 & 70.52 & 24.11 \\
    \bottomrule
  \end{tabular}
}

\vspace{2pt}
\emph{Note:} Friesian denotes FriesianCattle2017, SeaStar denotes SeaStarReID2023, and O2C denotes original-to-corrupted distillation. In Table~\ref{tab:router_ablation}, Fixed-2/Fixed-3 denote fixed averaging over two/three experts, and Top-2 denotes learned top-2 routing with the expert-pool size shown in parentheses.
\end{table}

Table~\ref{tab:avg_tradeoff} highlights the robustness--retrofit trade-off. Extra parameters are measured at inference relative to Std-FT, and gap recovery denotes the fraction of the CQ mAP gap between Std-FT and Corrupt-FT recovered by each method. DARA recovers 77.0\% of this gap with only 0.49\% additional inference parameters.

\subsubsection{Ablation Study}

\paragraph{Progressive component ablation.}
Table~\ref{tab:progressive_ablation} analyzes the contribution of residual adaptation and original-to-corrupted (O2C) distillation. One adapted layer gives the largest gain, raising average CQ mAP from 30.17 to 48.00, while the second layer provides a smaller complementary improvement to 50.31. O2C distillation further increases average CQ mAP to 55.70 and improves mINP, suggesting that the original-image teacher helps preserve harder identity matches under corruption. Standard mAP remains close to Std-FT, indicating that robustness does not require sacrificing unchanged-image retrieval.

\vspace{-4mm}
\paragraph{Effect of expert composition and learned routing.}
Table~\ref{tab:router_ablation} compares a single residual expert, fixed mixtures of multiple experts, and learned top-2 routing. Fixed mixtures improve over the single-expert variant, indicating that multiple correction patterns are useful. Learned routing improves further on all datasets, with the largest gain on MPDD, isolating the benefit of input-conditioned expert selection. Increasing the expert pool from three to four does not consistently improve performance, suggesting that adaptive routing is more important than simply adding expert capacity.

\vspace{-1mm}
\paragraph{Condition-dependent routing behavior.}

\begin{figure}[!t]
  \centering
  \IfFileExists{figures/router_weight_heatmap_4datasets.png}{%
    \includegraphics[width=\linewidth]
    {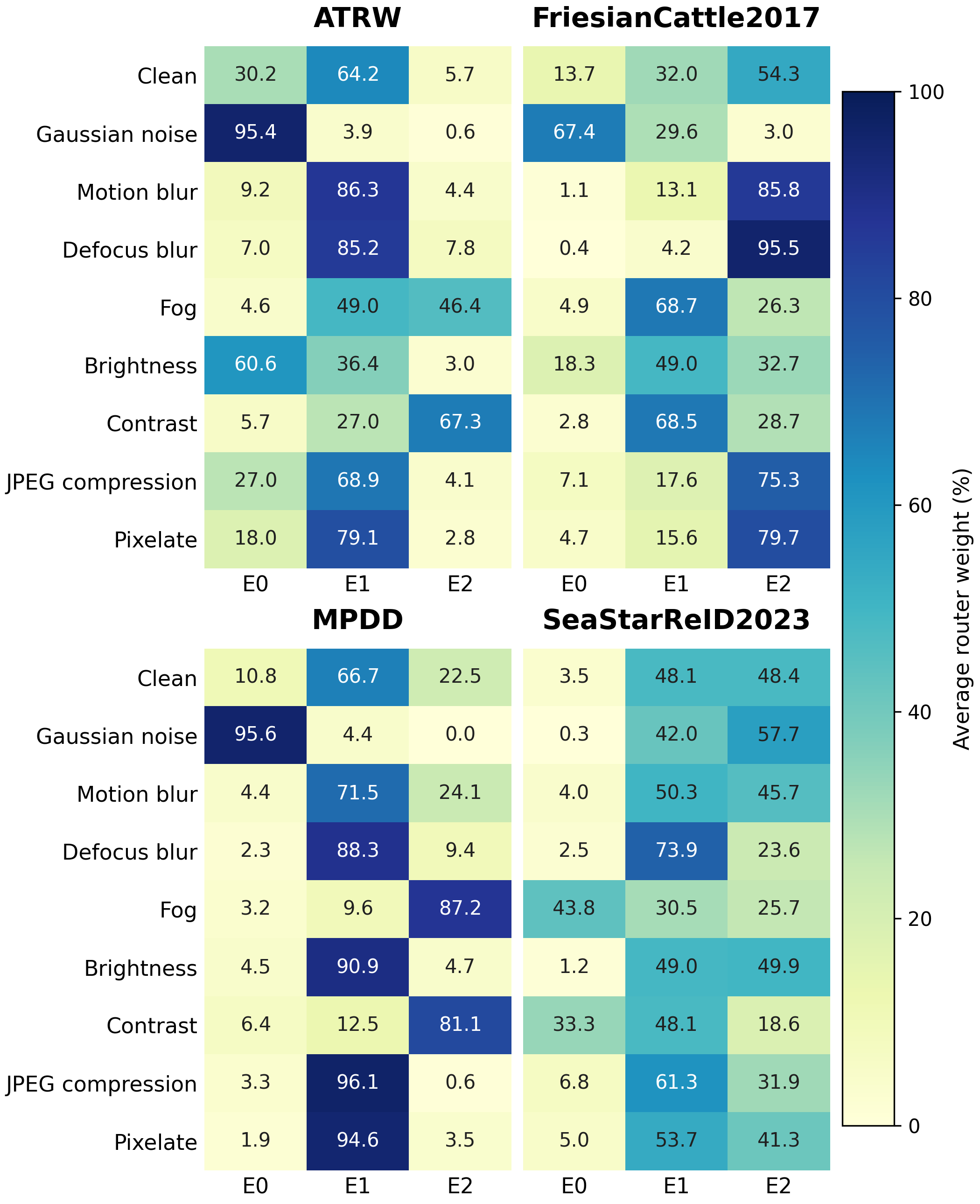}%
  }{%
    \fbox{%
      \parbox[c][0.18\textheight][c]{0.94\linewidth}{%
        \centering
        Placeholder for the four-dataset router-weight heatmap.
      }%
    }%
  }
  \caption{Condition-wise average routing weights of the soft top-2 router.
  Rows denote unchanged and corrupted input conditions, while columns denote the
  three residual experts.}
  \label{fig:router_weight_heatmap}
\end{figure}

Fig.~\ref{fig:router_weight_heatmap} visualizes condition-wise average routing
weights to examine whether expert selection changes with input degradation. The heatmap shows that expert weights vary across unchanged and corrupted conditions, with different experts emphasized for noise, blur, appearance, and digital corruptions. Since this behavior emerges without corruption labels, it suggests that DARA learns condition-dependent expert selection rather than fixed correction for every input.

\vspace{-1mm}
\subsubsection{Generalization Analysis}

\begin{figure}[!t]
  \centering
  \includegraphics[width=\linewidth]{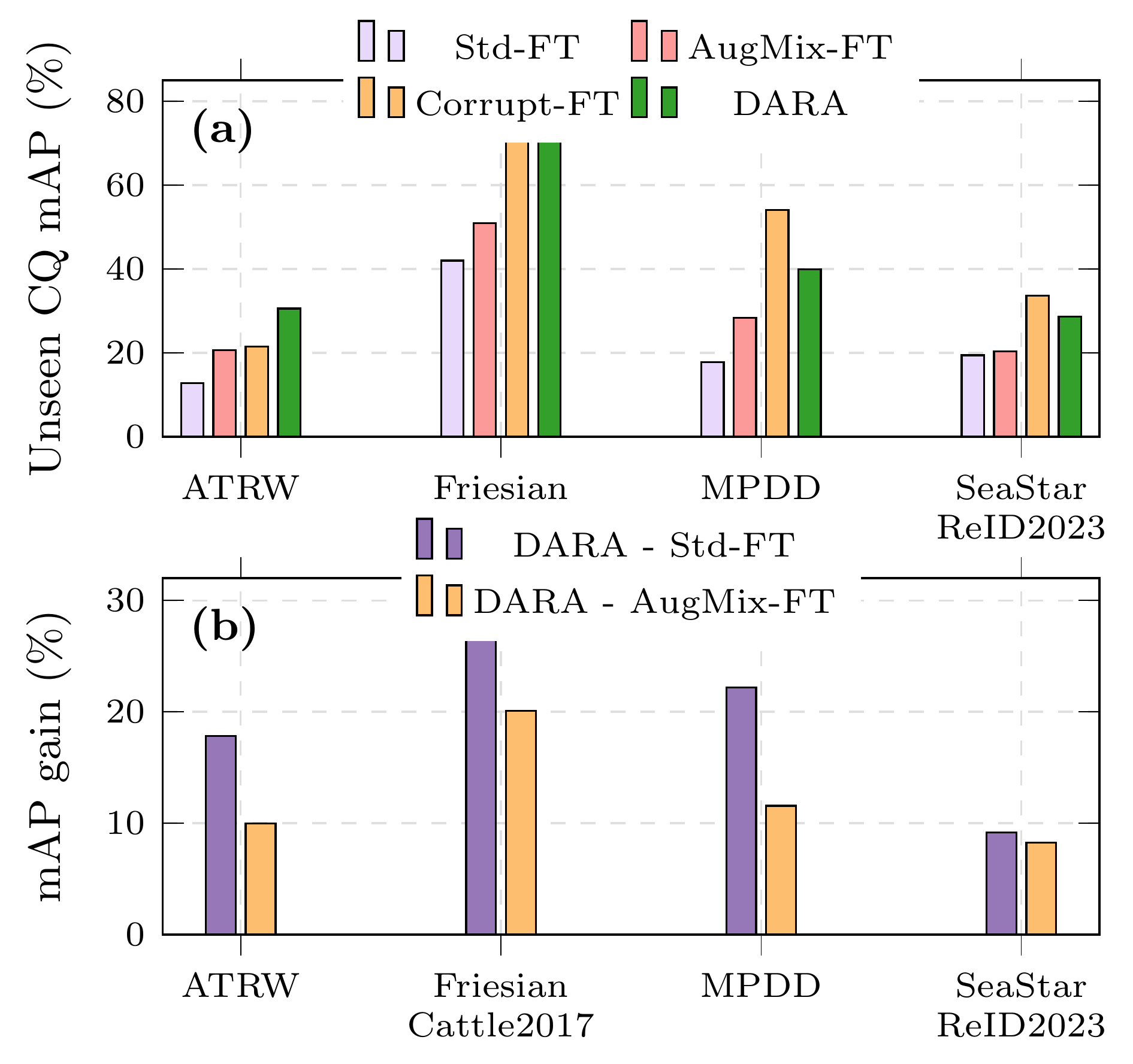}
  \caption{%
    Unseen corrupted-query evaluation.
    (a) Absolute mAP under unseen corrupted-query conditions across four animal
    Re-ID datasets.
    (b) DARA's CQ mAP gains over Std-FT and AugMix-FT.
  }
  \label{fig:unseen_cq}
\end{figure}

Fig.~\ref{fig:unseen_cq} evaluates corruptions not used during training. DARA improves CQ mAP over Std-FT and AugMix-FT on all four datasets, suggesting that the residual experts and router learn transferable degradation corrections rather than only fitting the training corruption set.

Beyond corruption-type shift, we also evaluate cross-domain CQ transfer from FriesianCattle2017 to CornwallCattle
\cite{chen2025animalreidmicrocontrollers}. DARA improves CQ mAP over Std-FT from 28.69\% to 39.31\%, showing that the learned repair transfers beyond the source dataset. Corrupt-FT remains higher at 45.92\%, which
is consistent with its use of full-backbone corrupted training. Full unseen-corruption and cross-domain CQ tables are provided in the supplementary.

\subsubsection{Deployment Efficiency}

At inference time, the teacher and distillation losses are removed, and only the adapted student is deployed. DARA adds 0.49\% parameters and 0.05\% FLOPs over Std-FT, with latency increases of 0.72\% on CPU and 6.87\% on H200. The H200 increase corresponds to only 0.23 ms, from 3.35 ms to 3.58 ms, showing that the robustness gain comes with little additional deployment cost.

%% file: sec/5_conclusion.tex
\vspace{-2mm}
\section{Conclusion}

Corrupted field images challenge animal Re-ID because degradation can distort the retrieval relationships required to match individuals across query and gallery images. We introduce DARA, a lightweight retrofit for corruption-robust animal Re-ID. DARA keeps the compact Re-ID backbone frozen and learns routed low-rank residual experts, guided by original-to-corrupted distillation, to repair degraded-input embeddings without corruption-type labels.

Experiments on four animal datasets show that DARA improves corrupted-query retrieval over standard and augmentation-based fine-tuning, while preserving unchanged-image performance. DARA also generalizes to unseen corruptions and cross-domain evaluation with limited parameter and computational overhead. These results suggest that corruption robustness in animal Re-ID can be recovered through input-conditioned feature repair, rather than relying only on full corrupted retraining of the backbone.
Future work will evaluate DARA under naturally occurring, mixed, and unlabeled degradations in the field.

%% file: sec/6_appendix.tex
\section{Additional Experimental Results}

\subsection{Dataset Statistics}
\label{app:dataset_stats}

\begin{table}[H]
\centering
\footnotesize
\caption{Statistics of the datasets used in our experiments.}
\label{tab:dataset}
\begin{tabular}{lcccccccc}
\toprule
Dataset & Train Images & Train IDs & Gallery Images & Gallery IDs & Query Images & Query IDs & Total Images & Total IDs \\
\midrule
ATRW & 3,730 & 149 & 521 & 33 & 424 & 33 & 4,675 & 182 \\
FriesianCattle2017 & 752 & 66 & 97 & 18 & 85 & 18 & 934 & 84 \\
MPDD & 1,032 & 95 & 521 & 96 & 104 & 96 & 1,657 & 191 \\
SeaStarReID2023 & 1,768 & 81 & 230 & 14 & 189 & 14 & 2,187 & 95 \\
\bottomrule
\end{tabular}
\end{table}
\FloatBarrier

\subsection{Full Unseen Corrupted-Query Results}
\label{app:unseen_cq}

\begin{table}[H]
  \centering
  \footnotesize
  \caption{
  Full unseen corrupted-query results. The table reports mAP, Rank-1, Rank-5,
  and mINP for all methods and datasets.
  }
  \label{tab:unseen_cq_full}
  \begin{tabular}{llccccccc}
    \toprule
    Dataset & Method
    & \multicolumn{3}{c}{Standard}
    & \multicolumn{4}{c}{Unseen CQ} \\
    \cmidrule(lr){3-5} \cmidrule(lr){6-9}
    & & mAP & R1 & R5 & mAP & R1 & R5 & mINP \\
    \midrule
    ATRW & Std-FT & 52.00 & 92.20 & 98.10 & 12.72 & 14.86 & 27.83 & 7.12 \\
    ATRW & AugMix-FT & 55.90 & 96.50 & 99.50 & 20.58 & 26.89 & 43.87 & 8.92 \\
    ATRW & Corrupt-FT & 52.40 & 94.60 & 99.10 & 21.46 & 35.85 & 58.02 & 7.62 \\
    ATRW & DARA & 54.20 & 94.30 & 98.80 & 30.54 & 47.41 & 63.92 & 12.06 \\
    \midrule
    FriesianCattle2017 & Std-FT & 95.10 & 98.80 & 100.00 & 42.00 & 36.47 & 55.29 & 35.57 \\
    FriesianCattle2017 & AugMix-FT & 97.10 & 100.00 & 100.00 & 50.92 & 45.88 & 61.18 & 45.81 \\
    FriesianCattle2017 & Corrupt-FT & 94.90 & 97.60 & 100.00 & 76.25 & 78.82 & 85.88 & 63.15 \\
    FriesianCattle2017 & DARA & 95.90 & 100.00 & 100.00 & 71.01 & 74.12 & 80.00 & 59.22 \\
    \midrule
    MPDD & Std-FT & 72.90 & 84.60 & 94.20 & 17.74 & 20.19 & 26.92 & 10.62 \\
    MPDD & AugMix-FT & 79.10 & 91.30 & 97.10 & 28.34 & 35.58 & 44.23 & 17.22 \\
    MPDD & Corrupt-FT & 74.70 & 88.50 & 98.10 & 54.03 & 61.54 & 75.00 & 36.07 \\
    MPDD & DARA & 73.20 & 85.60 & 93.30 & 39.91 & 50.96 & 62.50 & 23.01 \\
    \midrule
    SeaStarReID2023 & Std-FT & 55.20 & 69.30 & 93.10 & 19.43 & 13.76 & 28.04 & 14.01 \\
    SeaStarReID2023 & AugMix-FT & 67.10 & 81.50 & 95.20 & 20.35 & 18.52 & 38.10 & 13.68 \\
    SeaStarReID2023 & Corrupt-FT & 54.10 & 67.20 & 89.90 & 33.66 & 38.10 & 62.96 & 23.78 \\
    SeaStarReID2023 & DARA & 53.50 & 74.60 & 91.00 & 28.61 & 23.81 & 52.38 & 20.67 \\
    \bottomrule
  \end{tabular}
\end{table}
\FloatBarrier
\clearpage

\subsection{Additional Results under Different Corruption Locations}
\label{app:corruption_location}

\begin{table}[H]
  \centering
  \footnotesize
  \caption{Supplementary robustness results under different corruption
  locations, reported for completeness.}
  \label{tab:corruption_location_results}
  \begin{tabular}{llcccccccccccc}
    \toprule
    Dataset & Method
    & \multicolumn{4}{c}{Corrupted Eval}
    & \multicolumn{4}{c}{Corrupted Query}
    & \multicolumn{4}{c}{Corrupted Gallery} \\
    \cmidrule(lr){3-6} \cmidrule(lr){7-10} \cmidrule(lr){11-14}
    & & mAP & R1 & R5 & mINP & mAP & R1 & R5 & mINP & mAP & R1 & R5 & mINP \\
    \midrule
    ATRW & Std-ZS & 12.46 & 40.00 & 60.52 & 4.64 & 21.99 & 37.95 & 57.22 & 6.93 & 16.27 & 56.44 & 76.51 & 4.61 \\
    ATRW & Std-FT & 11.25 & 34.29 & 56.58 & 4.72 & 17.37 & 24.17 & 39.91 & 8.30 & 16.35 & 56.20 & 76.96 & 4.73 \\
    ATRW & AugMix-FT & 18.06 & 55.61 & 74.01 & 5.04 & 29.51 & 46.16 & 62.83 & 11.36 & 27.11 & 74.08 & 90.21 & 4.94 \\
    ATRW & Corrupt-FT & 45.21 & 86.58 & 95.97 & 13.16 & 48.82 & 90.09 & 96.96 & 16.65 & 47.74 & 91.20 & 97.48 & 13.70 \\
    ATRW & DARA & 34.58 & 71.53 & 89.72 & 10.23 & 43.28 & 75.33 & 88.96 & 15.91 & 41.52 & 84.13 & 96.39 & 11.64 \\
    \midrule
    FriesianCattle2017 & Std-ZS & 29.84 & 51.88 & 68.82 & 13.55 & 50.12 & 55.76 & 76.71 & 32.44 & 40.29 & 66.35 & 87.29 & 14.38 \\
    FriesianCattle2017 & Std-FT & 32.19 & 52.12 & 71.53 & 14.91 & 55.91 & 53.88 & 72.00 & 45.14 & 51.95 & 78.82 & 90.35 & 17.88 \\
    FriesianCattle2017 & AugMix-FT & 41.34 & 63.29 & 78.00 & 15.97 & 62.71 & 56.59 & 77.29 & 49.02 & 53.91 & 81.65 & 93.41 & 18.06 \\
    FriesianCattle2017 & Corrupt-FT & 87.62 & 94.94 & 99.06 & 67.81 & 90.57 & 96.47 & 99.88 & 77.99 & 90.04 & 95.88 & 99.06 & 72.28 \\
    FriesianCattle2017 & DARA & 75.48 & 87.53 & 97.41 & 48.00 & 85.07 & 90.12 & 97.65 & 72.17 & 83.75 & 96.24 & 99.65 & 52.76 \\
    \midrule
    MPDD & Std-ZS & 18.74 & 34.13 & 55.00 & 3.77 & 32.23 & 37.50 & 54.04 & 20.04 & 33.55 & 61.44 & 81.35 & 6.13 \\
    MPDD & Std-FT & 13.16 & 26.92 & 43.56 & 2.46 & 25.50 & 29.81 & 44.04 & 14.72 & 26.28 & 53.17 & 76.73 & 3.47 \\
    MPDD & AugMix-FT & 24.42 & 45.00 & 61.63 & 5.06 & 41.64 & 49.33 & 64.71 & 28.39 & 40.38 & 70.96 & 89.81 & 7.86 \\
    MPDD & Corrupt-FT & 60.54 & 77.40 & 92.60 & 35.88 & 68.25 & 82.88 & 94.81 & 47.23 & 67.34 & 85.87 & 96.35 & 41.73 \\
    MPDD & DARA & 40.87 & 58.17 & 79.71 & 16.24 & 56.92 & 68.85 & 82.79 & 36.88 & 53.81 & 75.96 & 89.04 & 21.82 \\
    \midrule
    SeaStarReID2023 & Std-ZS & 11.39 & 16.19 & 48.41 & 7.68 & 15.53 & 16.56 & 41.32 & 9.41 & 13.18 & 21.32 & 52.54 & 7.64 \\
    SeaStarReID2023 & Std-FT & 13.69 & 23.02 & 57.99 & 7.83 & 21.88 & 19.42 & 41.80 & 15.27 & 20.08 & 36.83 & 77.04 & 7.94 \\
    SeaStarReID2023 & AugMix-FT & 19.67 & 38.36 & 72.06 & 8.56 & 30.34 & 29.15 & 50.05 & 20.52 & 30.72 & 56.30 & 89.31 & 9.44 \\
    SeaStarReID2023 & Corrupt-FT & 41.27 & 51.38 & 87.46 & 23.91 & 45.76 & 55.08 & 83.49 & 29.82 & 45.53 & 56.72 & 89.26 & 25.10 \\
    SeaStarReID2023 & DARA & 28.87 & 36.67 & 77.41 & 16.70 & 37.53 & 41.90 & 73.39 & 24.93 & 34.34 & 47.04 & 84.23 & 18.30 \\
    \bottomrule
  \end{tabular}
\end{table}
\FloatBarrier

\subsection{Cross-Domain Corrupted-Query Results}
\label{app:cross_domain}

\begin{table}[H]
  \centering
  \footnotesize
  \caption{
  Full cross-domain corrupted-query results for models trained on
  FriesianCattle2017 and evaluated on CornwallCattle using seed 45.
  }
  \label{tab:cross_domain_cq_full}
  \begin{tabular}{lcccccc}
    \toprule
    Method
    & \multicolumn{3}{c}{Standard}
    & \multicolumn{3}{c}{Cross-Domain CQ} \\
    \cmidrule(lr){2-4} \cmidrule(lr){5-7}
    & mAP & R1 & R5 & mAP & R1 & R5 \\
    \midrule
    Std-ZS & 61.20 & 86.10 & 94.40 & 37.82 & 43.61 & 68.89 \\
    Std-FT & 54.10 & 75.00 & 97.20 & 28.69 & 28.61 & 62.22 \\
    Corrupt-FT & 57.90 & 72.20 & 97.20 & 45.92 & 55.83 & 82.22 \\
    DARA & 55.00 & 75.00 & 94.40 & 39.31 & 46.94 & 73.33 \\
    \bottomrule
  \end{tabular}
\end{table}
\FloatBarrier

\subsection{Deployment Efficiency}
\label{app:computational_efficiency}

\begin{table}[H]
  \centering
  \footnotesize
  \caption{Parameter and computational efficiency on CPU and GPU
  (FP32, batch size 64).}
  \label{tab:computational_efficiency}
  \begin{tabular}{lcccccccc}
    \toprule
    Method & Params & +Params & FLOPs & +FLOPs &
    CPU Lat. & +CPU Lat. & H200 Lat. & +H200 Lat. \\
    \midrule
    Std-FT & 3,150,336 & 0.00\% & 0.20501 & 0.00\% &
    866.39 & 0.00\% & 3.35 & 0.00\% \\
    DARA & 3,165,907 & +0.49\% & 0.20510 & +0.05\% &
    872.61 & +0.72\% & 3.58 & +6.87\% \\
    \bottomrule
  \end{tabular}
\end{table}